\documentclass[letterpaper]{article}
\usepackage{aaai18}
\usepackage{times}
\usepackage{helvet}
\usepackage{courier}
\usepackage{url}            
\usepackage{amsmath}
\usepackage{amssymb,amsmath,amsthm,enumitem}
\usepackage{helvet}
\usepackage{courier}
\usepackage{amssymb}
\usepackage{verbatim}
\usepackage{graphicx} 
\usepackage[ruled,vlined]{algorithm2e}
\usepackage{booktabs}
\usepackage{natbib}
\usepackage{multirow}
\usepackage{caption}

\newcommand{\beq}{\begin{equation}}
\newcommand{\eeq}{\end{equation}}
\newcommand{\beqs}{\begin{eqnarray}}
\newcommand{\eeqs}{\end{eqnarray}}
\newcommand{\barr}{\begin{array}}
	\newcommand{\earr}{\end{array}}

\newcommand{\etc}{\textit{etc.}}

\newcommand{\Amat}{{\bf A}}

\newcommand{\Cmat}{{\bf C}}

\newcommand{\Umat}{{\bf U}}
\newcommand{\Vmat}{{\bf V}}
\newcommand{\Wmat}{{\bf W}}
\newcommand{\Xmat}{{\bf X}}
\newcommand{\Ymat}{{\bf Y}}

\newcommand{\av}{{\boldsymbol a}}

\newcommand{\hv}{{\boldsymbol h}}

\newcommand{\mv}{{\boldsymbol m}}

\newcommand{\sv}{{\boldsymbol s}}

\newcommand{\wv}{{\boldsymbol w}}

\newcommand{\yv}{{\boldsymbol y}}
\newcommand{\zv}{{\boldsymbol z}}

\newcommand{\betav}{{\boldsymbol \beta}}

\newcommand{\R}{\mathbb{R}}
\newcommand{\E}{\mathbb{E}}

\newcommand{\Hcal}{\mathcal{H}}
\newcommand{\Lcal}{\mathcal{L}}
\newcommand{\Ncal}{\mathcal{N}}

\begin{document}
%
\title{Adaptive Feature Abstraction for Translating Video to Text}
\author{
	Yunchen Pu$^\dag$, Martin Renqiang Min$^\ddag$, Zhe Gan$^\dag$ and Lawrence Carin$^\dag$\\
	$^\dag$Department of Electrical and Computer Engineering, Duke University \\ \texttt{\{yunchen.pu, zhe.gan, lcarin\}@duke.edu} \\  
	$^\ddag$Machine Learning Group,	NEC Laboratories America \\
	\texttt{renqiang@nec-labs.com}
}
\maketitle
\begin{abstract}
Previous models for video captioning often use the output from a specific layer of a Convolutional Neural Network (CNN) as video features. However, the variable context-dependent semantics in the video may make it more appropriate to adaptively select features from the multiple CNN layers. We propose a new approach for generating adaptive spatiotemporal representations  of videos for the captioning task.  A novel attention mechanism is developed, that adaptively and sequentially focuses on different layers of CNN features (levels of feature ``abstraction''), as well as local spatiotemporal regions of the feature maps at each layer. The proposed approach is evaluated on three benchmark datasets: YouTube2Text, M-VAD and MSR-VTT.  Along with visualizing the results and how the model works, these experiments quantitatively demonstrate the effectiveness of the proposed adaptive spatiotemporal feature abstraction for translating videos to sentences with rich semantics.
\end{abstract}

\section{Introduction}
Videos represent among the most widely used forms of data, and their accurate characterization poses an important challenge for computer vision and machine learning. Generating a natural-language description of a video, termed video captioning, is an important component of video analysis. Inspired by the successful encoder-decoder framework used in machine translation~\citep{Bahdanau2015iclr,Cho2014emnlp,Sutskever2014nips} and image caption generation~\citep{neuraltalk,kiros2014multimodal,mao2015ICLR, Yunchen_NIPS,NIC,gan2017stylenet,zhe_cvpr}, most recent work on video captioning~\citep{donahue2015long,pan2016joint,venugopalan2015sequence,venugopalan2014translating,yao2015describing,yu2016video} employs a two-dimensional (2D) or three-dimensional (3D) Convolutional Neural Network (CNN) as an encoder, mapping an input video to a compact feature-vector representation. A Recurrent Neural Network (RNN) is typically employed as a decoder, unrolling the feature vector to generate a sequence of words of arbitrary length.

Despite achieving encouraging success in video captioning, previous models suffer important limitations. Often the rich video content is mapped to a single feature vector for caption generation; this approach is prone to missing detailed and localized spatiotemporal information. To mitigate this, one may employ methods to focus attention on local regions of the feature map, but typically this is done with features from a selected (usually top) CNN layer. By employing features from a fixed CNN layer, the algorithm is limited in its ability to model rich, context-aware semantics that requires focusing on different feature abstraction levels. As investigated in~\citet{mahendran2015understanding,Zeiler14ECCV}, the feature characteristics/abstraction is correlated with the CNN layers: features from layers at or near the top of a CNN tend to focus on global (extended) visual percepts, while features from lower CNN layers provide more local, fine-grained information. It is desirable to select/weight features from different CNN layers adaptively when decoding a caption, selecting different levels of feature abstraction by sequentially emphasizing features from different CNN layers. In addition to focusing on features from different CNN layers, it is also desirable to emphasize local spatiotemporal regions in feature maps at particular layers. 

To realize these desiderata, our proposed decoding process for generating a sequence of words dynamically emphasizes different levels (CNN layers) of 3D convolutional features, to model important coarse or fine-grained spatiotemporal structure. Additionally, the model adaptively attends to different locations within the feature maps at particular layers. While some previous models use 2D CNN features to generate video representations, our model adopts features from a deep 3D convolutional neural network (C3D). Such features have been shown to be effective for video representation, action recognition and scene understanding~\citep{tran2014learning}, by learning the spatiotemporal features that can provide better appearance and motion information. In addition, the proposed model is inspired by the recent success of attention-based models that mimic human perception~\citep{mnih2014recurrent,Attend}. 

The proposed model, dual Adaptive Feature Representation (dualAFR), involves comparing and evaluating different levels of 3D convolutional feature maps. However, there are three challenges that must be overcome to directly compare features between layers: (\emph{i}) the features from different C3D levels have distinct dimensions, undermining the direct use of a multi-layer perceptron (MLP) based attention model~\citep{Bahdanau2015iclr,Attend}; (\emph{ii}) the features represented in each layer are not spatiotemporally aligned,  undermining our ability to quantify the value of features at a specific spatiotemporal location, based on information from {\em all} CNN layers; and (\emph{iii}) the semantic meaning of feature vectors from the convolutional filters of C3D varies across layers, making it difficult to feed the features into the same RNN decoder. 

To address these issues, one may use either pooling or MLPs to map different levels of features to the similar semantic-space dimension. However, these approaches either lose feature information or have too many parameters to generalize well. In our approach, we employ convolution operations to achieve spatiotemporal alignment among C3D features from different levels and attention mechanisms to dynamically select context-dependent feature abstraction information.

The principal contributions of this paper are as follows: (\emph{i}) A new video-caption-generation model is developed by dynamically modeling context-dependent feature abstractions; 
(\emph{ii}) new attention mechanisms are developed to adaptively and sequentially emphasize different levels of feature abstraction (CNN layers), while also imposing attention within local spatiotemporal regions of the feature maps at each layer; 
(\emph{iii}) 3D convolutional transformations are introduced to achieve spatiotemporal and semantic feature consistency across different layers;
(\emph{iv}) the proposed model is demonstrated to outperform other multi-level feature based methods, such as hypercolumns~\citep{hypercolumns}, and achieves state-of-the-art performance on several benchmark datasets using diverse automatic metrics and human evaluations.

\section{Related Work}
Recent work often develops a probabilistic model of the caption, conditioned on a video.  In \citet{donahue2015long,venugopalan2015sequence,venugopalan2014translating,yu2016video,pan2015hierarchical} the authors performed video analysis by applying a 2D CNN pretrained on ImageNet, with the top-layer output of the CNN used as features.
Given the {\em sequence} of features extracted from the video frames, the video representation is then obtained by a CRF~\citep{donahue2015long}, mean pooling~\citep{venugopalan2014translating}, weighted mean pooling with attention~\citep{yu2016video}, or via the last hidden state of an RNN encoder~\citep{venugopalan2015sequence}. In~\citet{yao2015describing}, the 2D CNN features are replaced with a 3D CNN to model the short temporal
dynamics. These works were followed by~\citet{pan2016joint, Aalto}, which jointly embedded the 2D CNN features and spatiotemporal features extracted from a 3D CNN~\citep{tran2014learning}. More recently, there has been a desire to leverage auxiliary information to improve the performance of encoder-decoder models. In~\citet{MTVC}, auxiliary encoders and decoders are introduced to utilize extra video and sentence data. The entire model is learned by both predicting sentences conditioned on videos and self-reconstruction for the videos and captions. Similiar work includes~\citet{MMVD,TGM} where extra audio and topics of videos are leveraged.
However, all of these previous models utilize features extracted from the top layer of the CNN.

There is also work that combines mult-level features from a CNN. In~\citet{sermanet2013pedestrian}, a combination of intermediate layers is employed with the top layer for pedestrian detection, while a hypercolumn representation is utilized for object segmentation and localization in~\citet{hypercolumns} . Our proposed model is mostly related to~\citet{ballas2015delving}, but distinct in important ways. The intermediate convolutional feature maps are leveraged, like~\citet{ballas2015delving}, but an attention model is developed instead of the ``stack" RNN in~\citet{ballas2015delving}. In addition, a decoder enhanced with two attention mechanisms is constructed for generating captions, while a simple RNN decoder is employed in~\cite{ballas2015delving}. Finally,  we use features extracted from C3D instead of a 2D CNN.
\section{Method}\label{sec:method}
Consider $N$ training videos, the $n$th of which is denoted $\Xmat^{(n)}$, with associated caption $\Ymat^{(n)}$. The length-$T_n$ caption is represented $\Ymat^{(n)}=(\yv_1^{(n)},\dots,\yv^{(n)}_{T_n})$, with $\yv_t^{(n)}$ a 1-of-$V$ encoding vector, with $V$ the size of the vocabulary. 

For each video, the C3D feature extractor~\citep{tran2014learning} produces a set of features $\Amat^{(n)}=\{\av_1^{(n)},\dots,\av_L^{(n)},\av_{L+1}^{(n)}\}$, where $\{\av_1^{(n)},\dots,\av_L^{(n)}\}$ are feature maps extracted from $L$ convolutional layers, and $\av_{L+1}^{(n)}$ is a vector obtained from the top fully-connected layer. For notational simplicity, we omit all the superscript $(n)$ over {$\Xmat, \Ymat, \yv, \Amat, \av$} and subscript $n$ under $T$ throughout the paper.
Details are provided in Appendix.

\begin{figure*}[t]
	\includegraphics[width=0.95\textwidth]{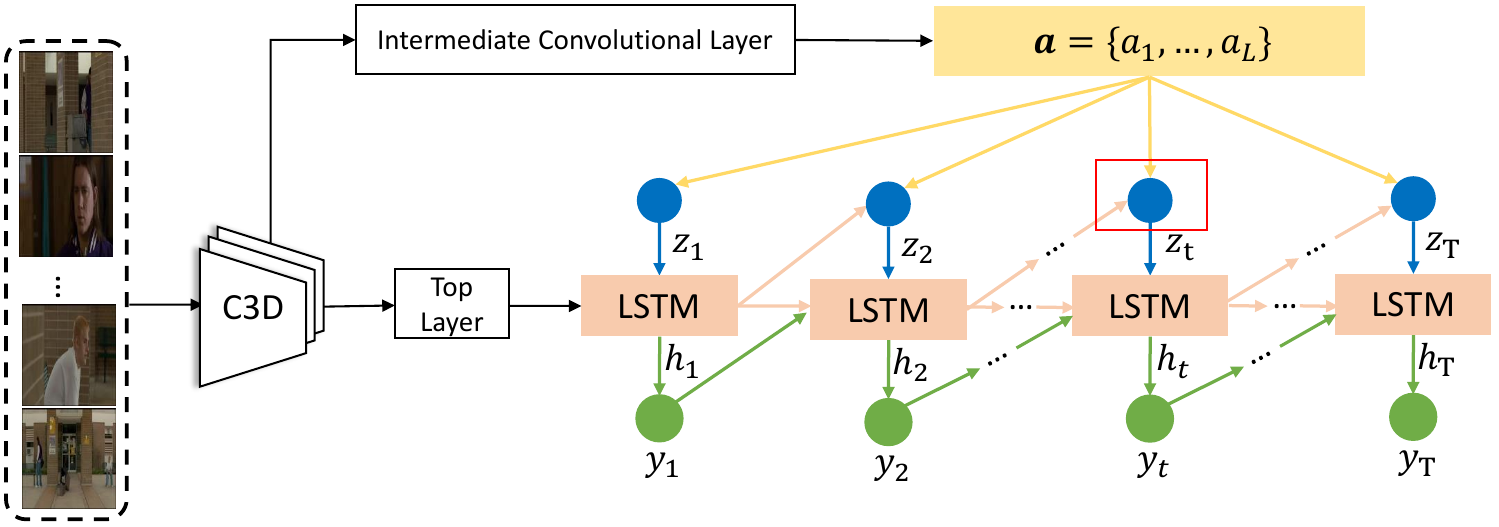}
	\caption{\small Illustration of our proposed caption-generation model. The model leverages a fully-connected map from the top layer as well as convolutional maps from different mid-level layers of a pretrained 3D  convolutional neural network (C3D). The context vector $\zv_t$ is generated from the previous hidden unit $\hv_{t-1}$ and the convolutional maps $\{\av_1,\dots,\av_L\}$ (the red frame), which is detailed in Figure~\ref{fig:attention}. }
	\vspace{-2mm}
	\label{fig:model}
\end{figure*}

\subsection{Caption Model}
The $t$-th word in a caption, $\yv_t$, is mapped to an $M$-dimensional vector $\wv_t=\Wmat_{e}\yv_t$, where $\Wmat_{e}\in \R^{M\times V}$ is a learned word-embedding matrix, \textit{i.e.}, $\wv_t$ is a column of $\Wmat_e$ chosen by the one-hot $\yv_t$. The probability of caption $\Ymat=\{\yv_t\}_{t=1,T}$ is defined as
\begin{align}
p(\Ymat|\Amat)=\textstyle p(\yv_1|\Amat){\prod_{t=2}^{T}p(\yv_t|\yv_{<t}, \Amat)}\,.\label{eq:decoder}
\end{align}
Specifically, the first word $\yv_1$ is drawn from $p(\yv_1|\Amat) = \mbox{softmax}(\Vmat \hv_1)$, where $\hv_1 = \tanh(\Cmat\av_{L+1})$; throughout the paper bias terms are omitted, for simplicity.  $\Cmat$ is a weight matrix mapping video features $\av_{L+1}$ to the RNN hidden state space.
$\Vmat$ is a matrix connecting the RNN hidden state to a softmax, for computing a distribution over words.
All other words in the caption are then sequentially generated using an RNN, until the end-sentence symbol is generated.
Conditional distribution $p(\yv_t|\yv_{<t},\Amat)$ is specified as $\mbox{softmax} (\Vmat \hv_t)$, where $\hv_t$ is recursively updated as $\hv_t = \Hcal(\wv_{t-1},\hv_{t-1},\zv_t)$.
$\zv_t = \phi(\hv_{t-1},\av_1,\dots,\av_L)$ is the context vector used in the attention mechanism, capturing the relevant visual features associated with the spatiotemporal attention (also weighting level of feature abstraction), as detailed below. 
The transition function $\Hcal(\cdot)$ is implemented with Long Short-Term Memory (LSTM)~\citep{hochreiter1997long}.

Given the video $\Xmat$ (with features $\Amat$) and associated caption $\Ymat$, the objective function is the sum of the log-likelihood of the caption conditioned on the video representation:
{\small
	\begin{align}
	\log p(\Ymat|\Amat) =\textstyle \log p(\yv_1|\Amat) +\sum_{t=2}^{T} \log p(\yv_t|\yv_{<t}, \Amat)\,.\label{eq:loss}
	\end{align}}
Equation (\ref{eq:loss}) is a function of all model parameters to be learned; they are not explicitly depicted in (\ref{eq:loss}) for notational simplicity. Further, (\ref{eq:loss}) corresponds to a single video-caption pair, and when training we sum over all such training pairs.  The proposed model is illustrated in Figure \ref{fig:model}.
\begin{figure*}[t!]
	\includegraphics[width=\textwidth]{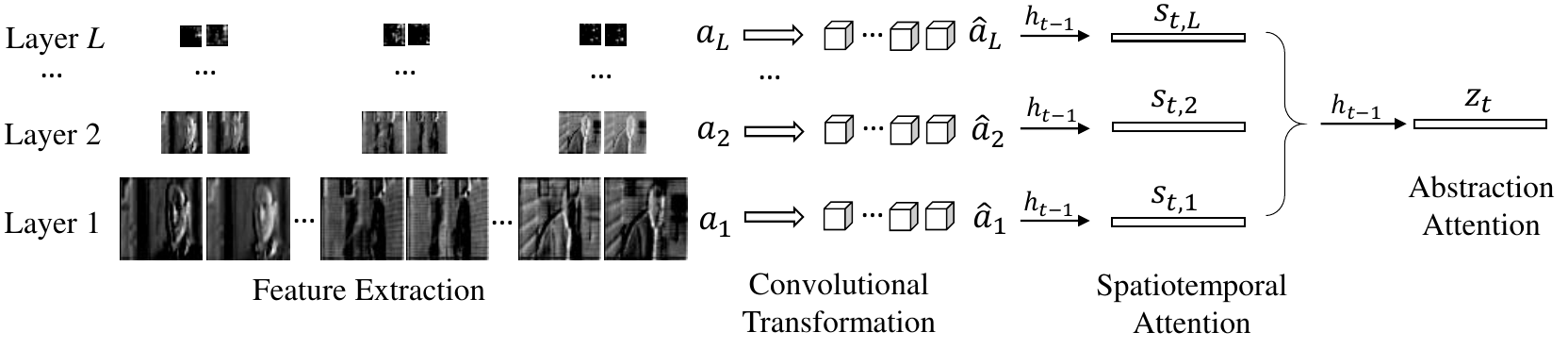}
	\caption{\small Illustration of our attention mechanism. The video features are extracted by C3D, followed by a 3D convolutional transformation. At each time step, the spatiotemporal attention takes these features and previous hidden units to generate $L$ feature vectors which are fed to the abstraction attention, manifesting the context vector. }
	\label{fig:attention}
	\vspace{-3mm}
\end{figure*}
\subsection{Attention Mechanism}\label{sec:att}
We first define notation needed to describe attention mechanism $\phi(\hv_{t-1},\av_1,\dots,\av_L)$. Each feature map, $\av_l$, is a 4D tensor, with elements corresponding to two spatial coordinates ($i.e.$, vertical and horizontal dimensions in a given frame), the third tensor index is the frame-index dimension, and the fourth dimension is associated with the filter index (for the convolutional filters). To be explicit, at CNN layer $l$, the number of dimensions of this tensor are denoted $n_x^l\times n_y^l\times n_z^l\times n_F^l$, with respective dimensions corresponding to vertical, horizontal, frame, and filter ($e.g.$, $n_F^l$ convolutional filters at layer $l$). Note that dimensions $n_x^l$, $n_y^l$ and $n_z^l$ vary with layer level $l$ (getting smaller with increasing $l$, due to pooling).

We define $\av_{i,l}$ as a ${n_F^l}$-dimensional {\em vector}, corresponding to a fixed coordinate in three of the tensor dimensions, $i.e.$, $i \in [1,\dots, n_x^l] \times [1,\dots, n_y^l] \times [1,\dots, n_z^l]$, while sweeping across all $n_F^l$ feature/filter dimensions. Further, define $\av_l(k)$ as a 3D tensor, associated with 4D tensor $\av_l$. Specifically, $\av_l(k)$ corresponds to the 3D tensor manifested from $\av_l$, with $k\in\{1,\dots,n_F^l\}$ a fixed coordinate in the dimension of the filter index. Hence, $\av_l(k)$ corresponds to the 3D feature map at layer $l$, due to the $k$th filter at that layer.

We introduce two attention mechanisms when predicting word $\yv_t$: ($i$) spatiotemporal-localization attention, and ($ii$) abstraction-level attention; these, respectively, measure the relative importance of a particular spatiotemporal location and a particular CNN layer (feature abstraction) for producing $\yv_t$, based on the word-history information $\yv_{<t}$. 

To achieve this, we seek to map $\av_l\rightarrow \hat{\av}_l$, 
where 4D tensors $\hat{\av}_l$ all have the same dimensions, are embedded into similar semantic spaces, and are aligned spatialtemporally. Specifically, $\hat{\av}_l$, $l=1,\dots,L-1$ are aligned in the above ways with $\av_L$. To achieve this, we filter each $\av_l$, $l=1,\dots, L-1$, and then apply max-pooling; the filters seek semantic alignment of the features (including feature dimension), and the pooling is used to spatiotemporally align the features with $\av_L$. Specifically, consider
\begin{align}
\hat{\av}_l&=\textstyle f(\sum_{k=1}^{n_F^l} \av_{l}(k)*\Umat_{k,l}),	\label{eq:conv_trans}	
\end{align} 
for $l=1,\dots,L-1$, and with $\hat{\av}_L = \av_L$. As discussed above, $\av_{l}(k)$ is the 3D feature map (tensor) for dictionary $k\in\{1,\dots,n_F^l\}$ at layer $l$. $\Umat_{k,l}$ is the 3D convolutional filters to achieve alignment which is a 4D learnable tensor. The convolution $*$ in (\ref{eq:conv_trans}) operates in the three shift dimensions, and $\av_{l}(k)*\Umat_{k,l}$ manifests a 4D tensor. Specifically, each of the $n_F^L$ 3D ``slices'' of $\Umat_{k,l}$ are spatiotemporally convolved (3D convolution) with $\av_{k,l}$, and after summing over the $n^l_F$ convolutional filters, followed by $f(\cdot)$, this manifests each of the $n^L_F$ 3D slices of $\hat{\av}_l$.
Function $f(\cdot)$ is an element-wise nonlinear activation function, followed by max pooling, with the pooling dimensions meant to realize final dimensions consistent with $\av_L$, $i.e.$, dimension $n_x^L \times n_y^L \times n_z^L \times n_F^L$. Consequently, $\hat{\av}_{i,l}\in \mathbb{R}^{ n_F^L}$ is a feature vector with $i\in [1,\dots, n_x^L] \times [1,\dots, n_y^L] \times [1,\dots, n_z^L]$. 
Note that it may only require one 3D tensor $\Umat_{l}$ applied on each 3D slices $\av_l(k)$ for each layer to achieve spatiotemporal alignment of the layer-dependent features.
However, the features from two distinct layers will not be in the same ``semantic'' space, making it difficult to assess the value of the layer-dependent features. The multiple tensors in set $\{\Umat_{k,l}\}$ provide the desired semantic alignment between layers, allowing analysis of the value of features from different layers via a single MLP, in the following \eqref{eq:spa_tem} and (\ref{eq:abs}).

With $\{\hat{\av}_{l}\}_{l=1,L}$ semantically and spatiotemporally aligned, we now seek to jointly quantify the value of a particular spatiotemporal region and a particular feature layer (``abstraction'') for prediction of the next word. 
For each $\hat{\av}_{i,l}$, the attention mechanism generates two positive weights, $\alpha_{ti}$ and $\beta_{tl}$, which measure the relative importance of location $i$ and layer $l$ for producing $\yv_t$ based on $\yv_{<t}$. Attention weights $\alpha_{ti}$ and $\beta_{tl}$ and context vector $\zv_{t}$ are computed as
\begin{align}
e_{ti}& = \wv_{\alpha}^T \tanh(\Wmat_{a\alpha}\hat{\av}_{i}+\Wmat_{h\alpha}\hv_{t-1}), \\
\alpha_{ti} &= \mbox{softmax}(\{e_{ti}\}),\quad	\sv_{t} = \psi(\{\hat{\av}_i\},\{\alpha_{ti}\}),\label{eq:spa_tem}  \\
b_{tl}& = \wv_{\beta}^T \tanh(\Wmat_{s\beta}\sv_{tl}+\Wmat_{h\beta}\hv_{t-1}), \\
\beta_{tl} &= \mbox{softmax}(\{b_{tl}\}),\quad	\zv_{t} = \psi(\{\sv_{tl}\},\{\beta_{tl}\}),\label{eq:abs}
\end{align}
where $\hat{\av}_i$ is a vector composed by stacking $\{\hat{\av}_{i,l}\}_{l=1,L}$ (all features at position $i$). $e_{ti}$ and $b_{tl}$ are scalars reflecting the importance of spatiotemporal region $i$ and layer $l$ to predicting $\yv_t$, while $\alpha_{ti}$ and $\beta_{tl}$ are {\em relative} weights of this importance, reflected by the softmax output. $\psi(\cdot)$ is a function developed further below, that returns a single feature vector when given a set of feature vectors, and their corresponding weights across all $i$ or $l$. Vector $\sv_{tl}$ reflects the sub-portion of $\sv_t$ associated with layer $l$.

In (\ref{eq:spa_tem}) we provide attention in the spatiotemporal dimensions, with that spatiotemporal attention shared across all $L$ (now aligned) CNN layers. In (\ref{eq:abs}) the attention is further refined, focusing attention in the layer dimension.
To make the following discussion concrete, we describe the attention function within the context of $\zv_{t} = \psi(\{\sv_{tl}\},\{\beta_{tl}\})$. This function setup is applied in the same way to $\sv_{t} = \psi(\{\hat{\av}_i\},\{\alpha_{ti}\})$. The attention model is illustrated in Figure~\ref{fig:attention}.
\vspace{-15pt}
\paragraph{Soft attention}
Following~\citet{Bahdanau2015iclr}, we formulate the soft attention model by computing a weighted sum of the input features
\begin{align}
\zv_{t} = \psi(\{\sv_{tl}\},\{\beta_{tl}\})= \textstyle \sum_{l=1}^{L} \beta_{tl} \sv_{tl}.\label{eq:soft}
\end{align}
The model is differentiable for all parameters and can be learned end-to-end using standard back-propagation.
\paragraph{Hard attention}
Let $\mv_t\in\{0,1\}^{L}$ be a vector of all zeros, and a single one, and the location of the non-zero element of $\mv_t$ identifies the location to extract features for generating the next word. We impose
\begin{align}
\mv_t \sim \mathsf{Mult}(1,\{\beta_{tl}\}), \qquad\zv_{t} =\textstyle\sum_{l=1}^{L} m_{tl}\sv_{tl}.
\end{align}
In this case, optimizing the objective function in \eqref{eq:loss} is intractable. However, the marginal log-likelihood can be lower-bounded as
\begin{align} \label{eq:lbound}
\log p(\Ymat|\Amat) &= \E_{p(\mv|\Amat)}\log p(\Ymat|\mv,\Amat),
\end{align}
where $\mv = \{\mv_t\}_{t=1,\dots,T}$. We utilize Monte Carlo integration to approximate the expectation, $\E_{p(\mv|\Amat)}$, and stochastic gradient descent (SGD) for parameter optimization. The gradients are approximated by importance sampling~\citep{mnih2016variational, Burda16importance} to achieve unbiased estimation and reduce the variance. Details are provided in Appendix.

\section{Experiments}
\subsection{Datasets}
We present results on three benchmark datasets: Microsoft Research Video Description Corpus (YouTube2Text)~\citep{Y2T}, Montreal Video Annotation Dataset (M-VAD)~\citep{MVAD}, and Microsoft Research Video to Text (MSR-VTT)~\citep{MSRVTT}. 

The Youtube2Text contains 1970 Youtube clips, and each video is annotated with around 40 sentences. For fair comparison, we used the same splits as provided in~\citet{venugopalan2014translating}, with 1200 videos for training, 100 videos for validation, and 670 videos for testing. 

The M-VAD is a large-scale movie description dataset, which is composed of 46587 movie snippets annotated with 56752 sentences. We follow the setting in~\cite{MVAD}, taking 36920 videos for training, 4950 videos for validation, and 4717 videos for testing. 

The MSR-VTT is a newly collected large-scale video dataset, consisting of 20 video categories. The dataset was split into 6513, 2990 and 497 clips in the training, testing and validation sets. Each video has about 20 sentence descriptions. The ground-truth captions in the testing set are not available now. Thus, we split the original training dataset into a training set of 5513 clips and a testing set of 1000 clips.

We convert all captions to lower case and remove the punctuation, yielding vocabulary sizes $V=12594$ for Youtube2Text, $V=13276$ for M-VAD, and $V=8071$ for MSR-VTT. 

We consider the RGB frames of videos as input, and all videos are resized to $112 \times 112$ spatially, with 2 frames per second. 
The C3D~\citep{tran2014learning} is pretrained on Sports-1M dataset~\citep{karpathy2014large}, consisting of 1.1 million sports videos belonging to 487 categories. We extract the features from four convolutional layers and one fully connected layer, named as  {\em pool2, pool3, pool4, pool5} and {\em fc-7} in the C3D~\citep{tran2014learning}, respectively. 
Detailed model architectures and training settings are provided in Appendix.
\begin{table*}[t!]
	\centering
	\caption{\small Results on BLEU-4, METEOR and CIDEr metrics compared to other models and baselines on Youtube2Text and M-VAD datasets.}\label{tab:results1}
	\vspace{-2mm}
	\begin{small}
		\begin{tabular}{c|ccc|ccc}
			\toprule
			\multirow{2}{*}{Methods}& \multicolumn{3}{c|}{Youtube2Text} & \multicolumn{3}{c}{M-VAD}\\
			\cline{2-7}
			& BLEU-4 & METEOR & CIDEr & BLEU-4 & METEOR & CIDEr \\
			\midrule
			LSTM-E (VGG + C3D)~\citep{pan2016joint} & 45.3 & 31.0 & - &- & 6.7 &-\\				
			GRU-RCN~\citep{ballas2015delving}  & 47.90 &  31.14 & 67.82 & - & - & -\\
			h-RNN (C3D+VGG)~\citep{yu2016video} & 49.9 & 32.6 & 65.8 & - & - & -\\
			TGM~\citep{TGM}&48.76 &34.36 &80.45 &- &- &-\\
			M-to-M~\citep{MTVC} & {\bf 54.5} & 36.0 & {\bf 92.4} & - & 7.4 & - \\
			
			\midrule
			\textit{Baselines} & \multicolumn{6}{c}{ }\\			
			\midrule
			ResNet + LSTM & 44.08 & 30.99 & 66.88 & 0.81  & 6.22 & 5.54\\
			C3D fc7 + LSTM & 45.34 & 31.21 & 66.12 & 0.83 & 6.31 & 5.96 \\
			C3D fc7 + pool2 & 45.46 & 31.53 & 67.38 & 0.98 & 6.42 & 6.01\\
			C3D fc7 + pool3 & 48.07 & 33.52 & 69.97 & 1.12 & 6.71 & 6.97\\
			C3D fc7 + pool4 & 48.18 & 34.47 & 70.97 & 1.24 & 6.89 & 7.12\\
			C3D fc7 + pool5 & 47.75 & 33.35 & 69.71 & 1.02 & 6.49 & 6.48\\
			\midrule
			\textit{Baselines of multi-level features with attention } & \multicolumn{6}{c}{ }\\
			\midrule			
			MLP + soft attention& 35.99 & 23.56 & 44.21 & 0.41 & 5.99 & 5.43\\
			Max-pooling + soft attention& 45.35 & 31.50 & 62.99 & 0.82 & 6.21 & 5.55\\
			Average-pooling + soft attention& 48.53 & 33.59 & 65.07 & 0.91 & 6.45 & 6.12\\
			Hypercolumn + soft attention & 44.92 & 30.18 & 63.18 & 0.88 & 6.33 & 6.24\\			
			\midrule
			\textit{dualAFR} & \multicolumn{6}{c}{ }\\
			\midrule		
			Soft Attention (Single model)  & 51.74 & 36.39  & 72.18 & 1.94 &  7.72 & 7.98\\
			Hard Attention (Single model)  & 51.77 & 36.41  &  72.21 & 1.82 & 7.12  &  8.12\\
			Soft Attention (Ensemble of 10)  & 53.94 & 37.91  & 78.43 & {\bf 2.14} & {\bf 8.22} & 9.03\\
			Hard Attention (Ensemble of 10)  & 54.27 & {\bf 38.03}  &  78.31 & 2.08 & 7.12  & {\bf 9.14}\\			
			\bottomrule
		\end{tabular}
		\vspace{-3mm}
	\end{small}
\end{table*}
We do not perform any dataset-specific tuning and regularization other than dropout~\citep{dropout} and early stopping on validation sets.
\subsection{Evaluation}
The widely used BLEU~\citep{bleu}, METEOR~\citep{meteor} and CIDEr~\citep{cider} metrics are employed to
quantitatively evaluate the performance of our video caption generation model, and other models in the literature. 
For each dataset, we show three types of results, using part of or all of our model to illustrate the role of each component: 
\begin{enumerate}
	\item
	{\em C3D fc7 + LSTM }: The LSTM caption decoder is employed, only using features extracted from the top fully-connected layer. No context vector $\zv_t$ is generated from intermediate convolutional layer features.
	\item
	{\em Spatiotemporal attention + LSTM}: The context vector $\zv_t$ is included, but only features extracted from a certain convolutional layer are employed, \textit{i.e.}, $\zv_t$ is equal to $\sv_t$ in \eqref{eq:spa_tem}. The spatiotemporal attention is implemented with the soft attention in \eqref{eq:soft}. 
	\item
	{\em dualAFR}: This is our proposed model. We present results based on both soft attention and hard attention.
\end{enumerate}
To compare our method with other multi-level feature methods, we show several baseline results on Youtube2Text:
\begin{figure*}[th!]
	\centering
	\includegraphics[width=0.9\textwidth]{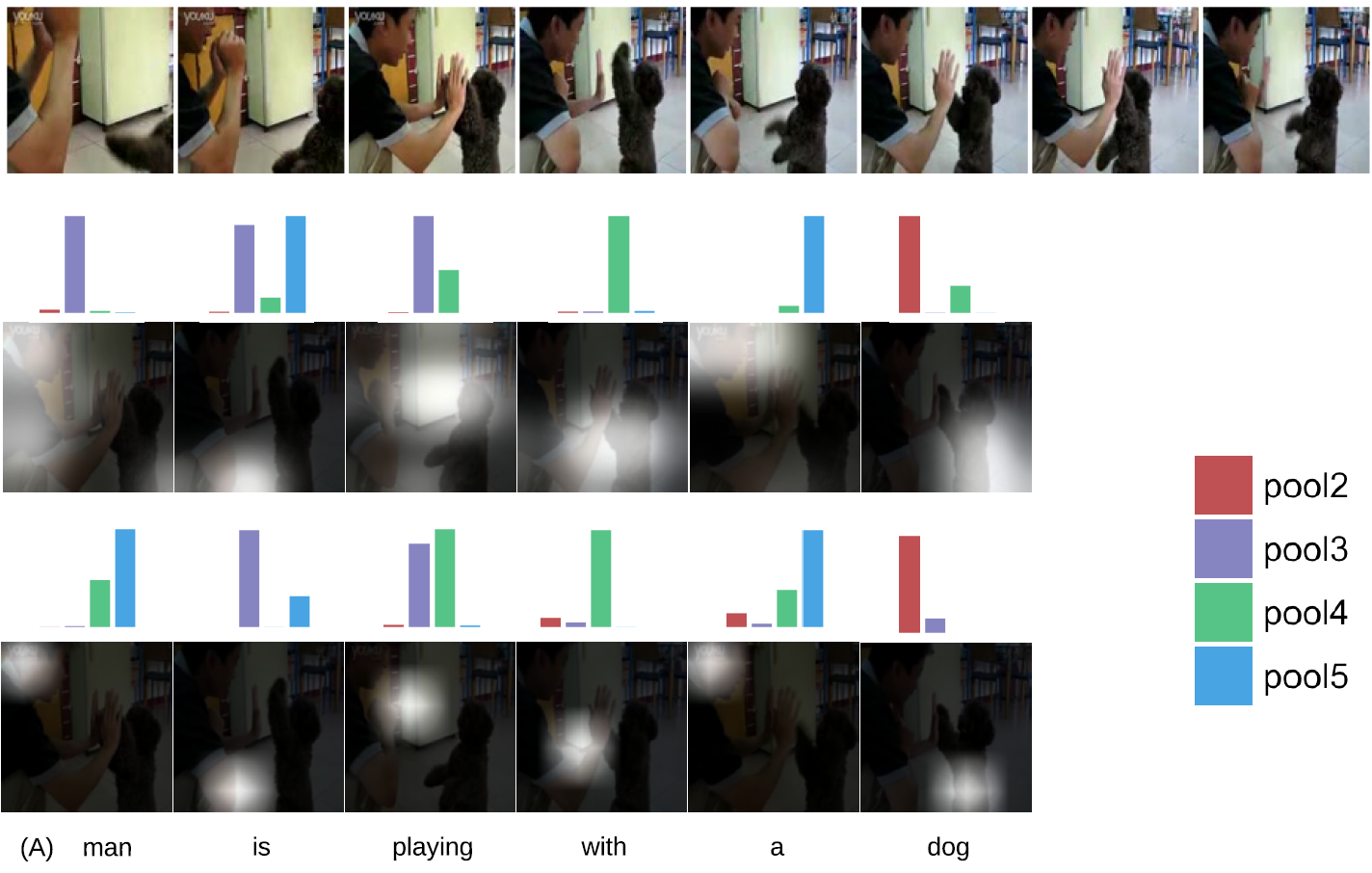}
	\caption{\small Visualization of attention weights aligned with input video and generated caption on Youtube2Text: (top) Sampled frames of input video, (middle) soft attention, (bottom) hard attention. We show the frame with the strongest attention weights. The bar plot above each frame corresponds layer attention weights $\betav_{tl}$ when the corresponding word (under the frame) is generated.} 
	\label{fig:vis_att}
\end{figure*}
\begin{enumerate}
	\item
	{\em MLP }: Each $\av_l$ is fed to a different MLP to achieve the same dimension of $\av_L$. 
	The context vector $\zv_t$ is obtained by abstraction-level and spatiotemporal attention.
	\item
	{\em Max/Average-pooling}:  A max or average pooling operation is utilized to achieve saptiotemporal alignment, and an MLP is then employed to embed the feature vectors into the similar semantic space. Details are provided in Appendix.
	\item
	{\em Hypercolumn}: This method is similar to max/average-pooling but replace the pooling operation with hypercolumn representation~\citep{hypercolumns}. 
\end{enumerate}
In these methods, the attention weights are produced by the soft attention in \eqref{eq:soft}.

To verify the effectiveness of our video caption generation model and C3D features, we also implement a strong baseline method based on the LSTM encoder-decoder network~\citep{Cho2014emnlp}, where ResNet~\citep{resnet} is employed as the feature extractor on each frame. We denote results using this method as {\em ResNet + LSTM }.

We also present human evaluation results based on relevance 
and coherence
\footnote{Our human evaluation follows the algorithm in the COCO captioning challenge (http://cocodataset.org/dataset.htm\#captions-challenge2015).}. Compared with single layer baseline models, the inference of dualAFR is about $1.2\sim 1.5$ times slower and requires about $2\sim4$ times extra memory.
\begin{figure*}[th!]
	\centering
	\includegraphics[width=0.9\textwidth]{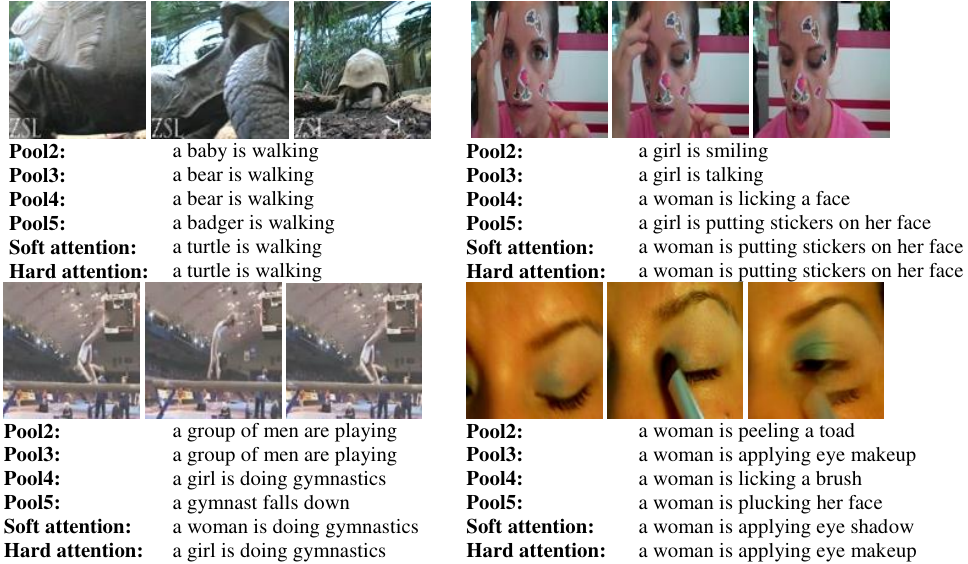}
	\caption{\small Examples of generated captions with sampled frames of input video on YouTube2Text.} 
	\label{fig:gen_cap}
\end{figure*}
\begin{table}[t!]
	\centering
	\caption{\small{Results on BLEU-4, METEOR and CIDEr metrics compared to other models and baselines on MSR-VTT.} $^\S$\citep{MMVD}; $^\dagger$\citep{Aalto}; $^\mathsection$\citep{TGM}; $^\ddagger$\citep{MTVC}.}\label{tab:results2}
	\vspace{-2mm}
	\begin{small}
		\begin{tabular}{c|ccc}
			\toprule
			Method & BLEU-4 & METEOR & CIDEr\\
			\midrule
			MMVD$^\S$ & 40.7& 28.6& 46.5\\
			M-to-M$^\ddagger$& 40.8 &28.8 &47.1 \\
			Aalto$^\dagger$	& 41.1&  27.7&  46.4 \\
			TGM$^\mathsection$ & 44.33 &29.37 &49.26\\
			\midrule
			\textit{Baselines} & \multicolumn{3}{c}{ }\\				
			\midrule
			ResNet + LSTM & 39.54 & 26.59 & 45.22 \\
			C3D fc7 + LSTM & 40.17 & 26.86 & 45.95 \\
			C3D fc7 +  pool2 & 40.43 & 26.93 & 47.15\\
			C3D fc7 + pool3 & 42.04 & 27.18 & 48.93\\
			C3D fc7 +  pool4 & 41.98 & 27.42 & 48.21\\
			C3D fc7 + pool5 & 40.83 & 27.01 & 47.86\\
			\midrule
			\textit{dualAFR} & \multicolumn{3}{c}{ }\\
			\midrule
			Soft Attention (Single) &  43.72 & 29.67  &  50.21\\
			Hard Attention (Single) &  43.89 &  28.71  &  50.29\\
			Soft Attention (Ensemble) &  44.99 & {\bf 30.16}  &  51.13\\
			Hard Attention (Ensemble) &  {\bf 45.01} &  29.98  &  {\bf 51.41}\\
			\bottomrule
		\end{tabular}
		\vspace{-1mm}
	\end{small}
\end{table}
\subsection{Quantitative Results}
Results are summarized in Tables \ref{tab:results1} and \ref{tab:results2}. The proposed models achieve state-of-the-art results on most metrics on all three datasets. The M-to-M method~\citep{MTVC} is the only model showing better BLEU and CIDEr on Youtube2Text. Note that the M-to-M model is trained with two additional datasets: UFC-101 which contains 13,320 video clips and Stanford Natural Language Inference corpus which contains 190,113 sentence pairs. 
In contrast, we achieve competitive or better results by using only the data inside the training set and analogous pretrained C3D.

Note that our model consistently yields significant improvements over models with only spatiotemporal attention, which further achieve better performance than only taking the C3D top fully-connected layer features; this demonstrates the importance of leveraging intermediate convolutional layer features. In addition, our model outperforms all the baseline results of multi-level feature based methods, demonstrating the effectiveness of our 3D convolutional transformation operation. This is partly a consequence of the sparse connectivity of the convolution operation, which indicates fewer parameters are required.

\subsection{Human Evaluation}
Besides the automatic metrics, we present human evaluation on the Youtube2Text dataset. In each survey, we compare our results from single model (soft attention or hard attention) with the strongest baseline ``C3D fc3 + pool4" by taking a  random sample of 100 generated captions, and ask the human evaluator to select the result with better relevance and coherence. We obtain 25 repsonses (2448 samples in total) and the results are shown in Table~\ref{tab:human}. The proposed dualAFR outperforms the strongest baseline on both relevence and coherence which is consistent with the automatic metrics.
\subsection{Qualitative Results}
Following~\cite{Attend}, we visualize the attention components learned by our model on Youtube2Text. As observed from Figure \ref{fig:vis_att}, the spatiotemporal attention aligns the objects in the video well with the corresponding words. In addition, the abstraction-level attention tends to focusing on low level features when the model generates a noun and high level features when an article or a preposition is being generated. More results are provided in Appendix.
Examples of generated captions from unseen videos on Youtube2Text are shown in Figure \ref{fig:gen_cap}. We find the results with abstraction-layer attention (indicated as ``soft attention" or `` hard attention") is generally equal to or better than the best results, compared to those only taking a certain convolutional-layer feature (indicated as ``Pool2" \etc). This demonstrates the effectiveness of our abstraction layer attention. 
More results are provided in Appendix.
\begin{table}[t!]
	\centering
	\caption{\small{Human evaluation results on Youtube2Text.}}\label{tab:human}
	\vspace{-2mm}
	\begin{small}
		\begin{tabular}{c|cc}
			\toprule
			Method & Relevance & Coherence \\
			\midrule
			Baseline wins & 4.78\% &  1.63\% \\
			DualAFR wins &  27.53\% &6.49\% \\
			Not distinguishable & 67.69\% &91.88\% \\
			\bottomrule
		\end{tabular}
	\end{small}
\end{table}

\section{Conclusion and Future Work}
A novel video captioning model has been proposed, that adaptively selects/weights the feature abstraction (CNN layer), as well as the location within a layer-dependent feature map. We have implemented the attention using both ``hard'' and ``soft'' mechanisms, with the latter typically delivering better performance. Our model achieves excellent video caption generation performance, and has the capacity to provide interpretable alignments {\em seemingly} analogous to human perception.

We have focused on analysis of videos and associated captions.
Similar ideas may be applied in the future to image captioning. Additionally, the CNN parameters were learned separately as a first step, prior to analysis of the captions. It is also of interest to consider CNN-parameter refinement conditioned on observed training captions.
\small{
\bibliographystyle{aaai}
\bibliography{references}
}

\twocolumn[{%
	\renewcommand\twocolumn[1][]{#1}%
	\maketitle
	\appendix
	\section{More results}
	\subsection{Visualization of attention weights}
	\begin{center}
		\centering
		\includegraphics[width=0.82\textwidth]{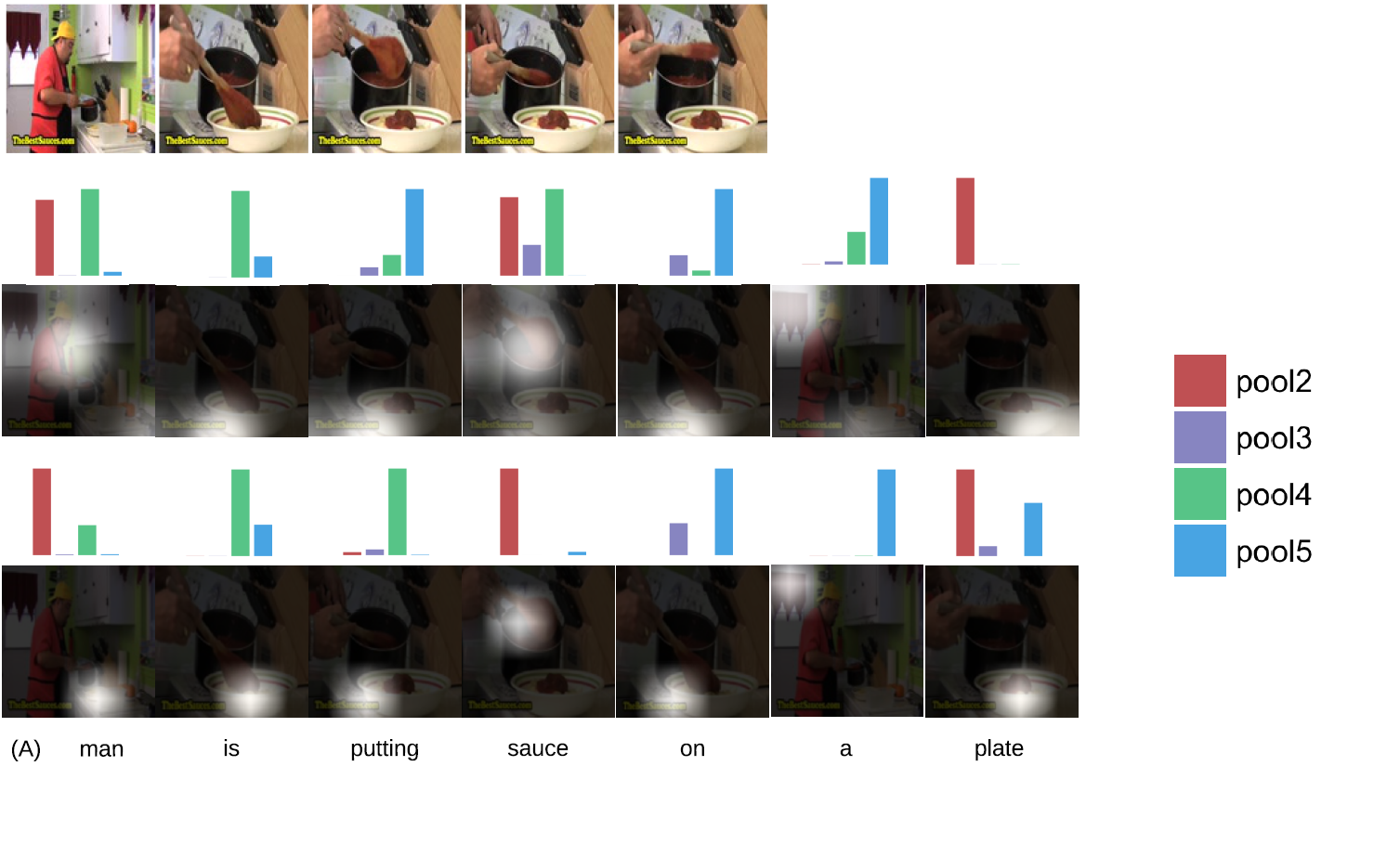}\\
		\includegraphics[width=0.82\textwidth]{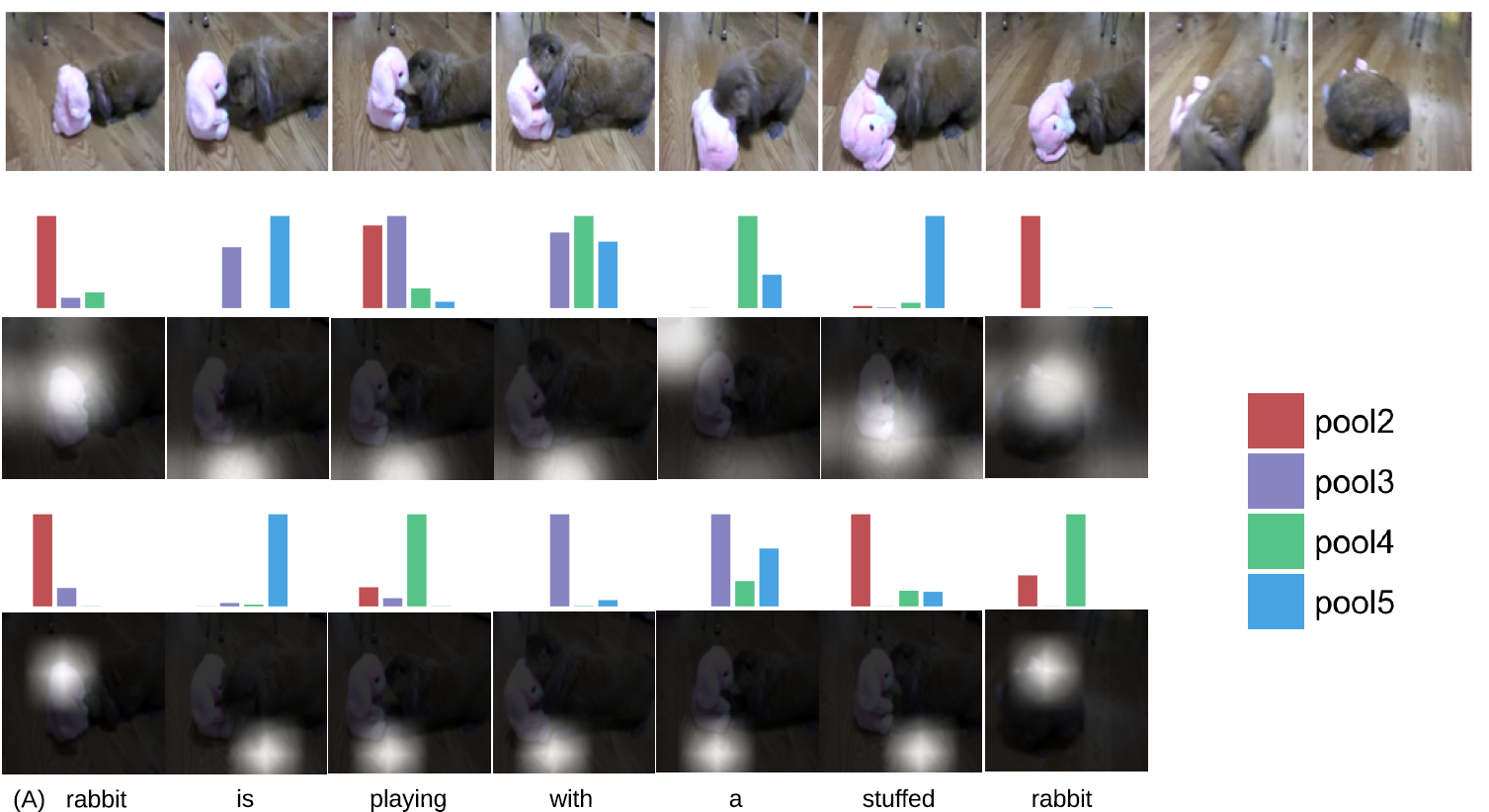}
	\end{center}
}]
\clearpage
\begin{figure*}[h]
	\centering
	\includegraphics[width=0.82\textwidth]{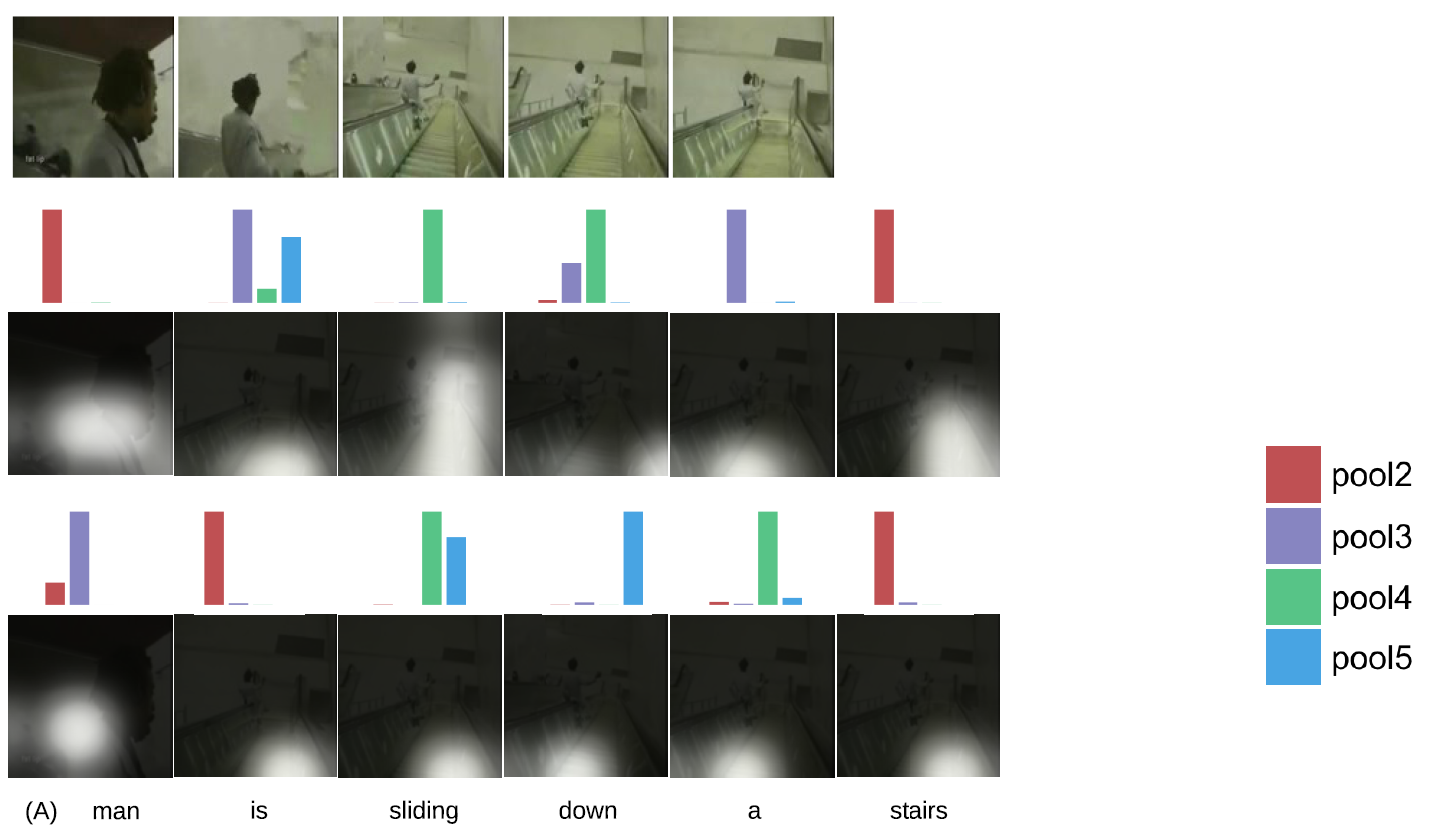}\\
	\vspace{30mm}
	\includegraphics[width=0.82\textwidth]{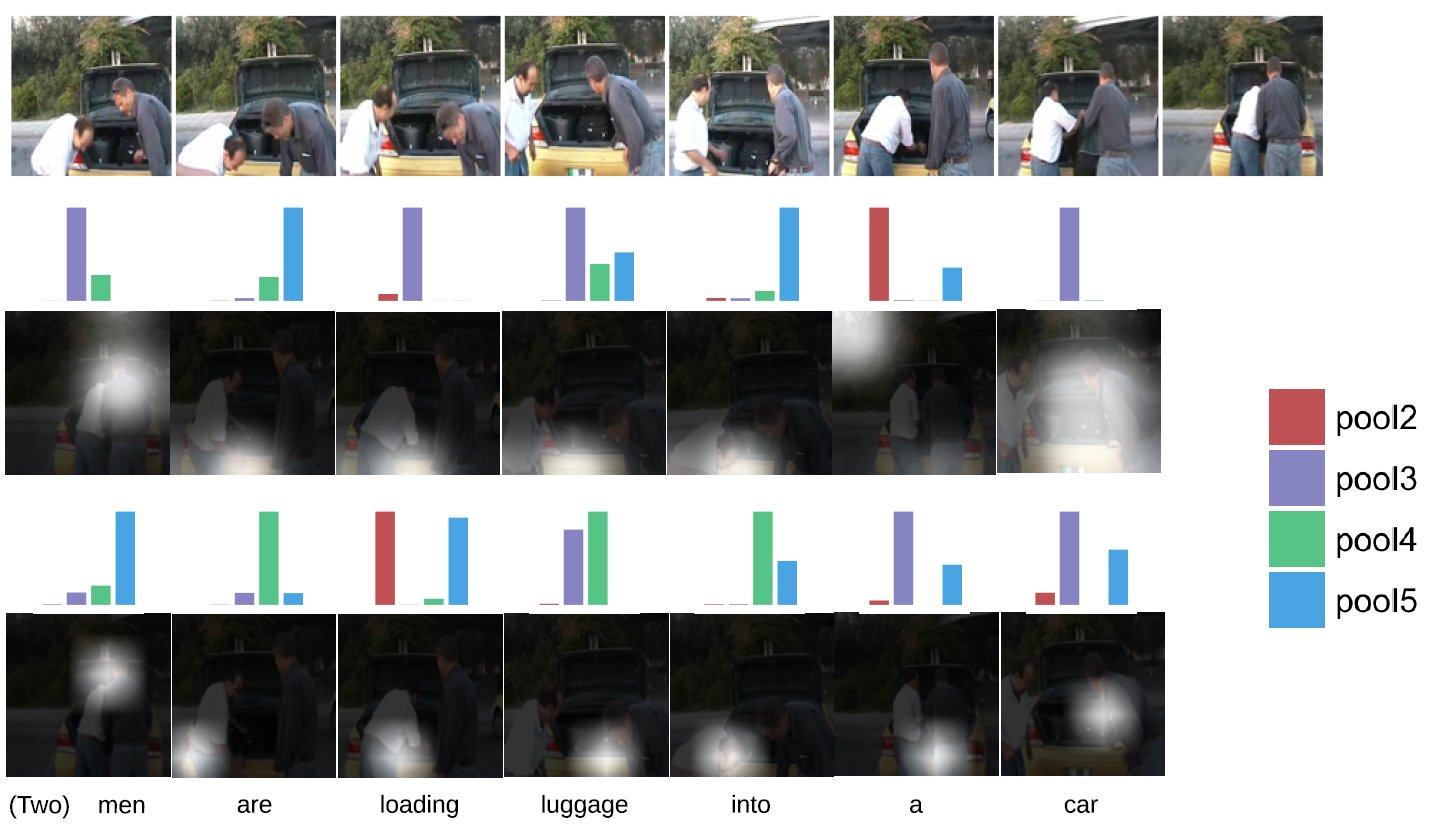}
\end{figure*}

\clearpage
\twocolumn[{%
	\renewcommand\twocolumn[1][]{#1}%
	\subsection{Generated Captions}
	\begin{center}
		\centering
		\includegraphics[width=\textwidth]{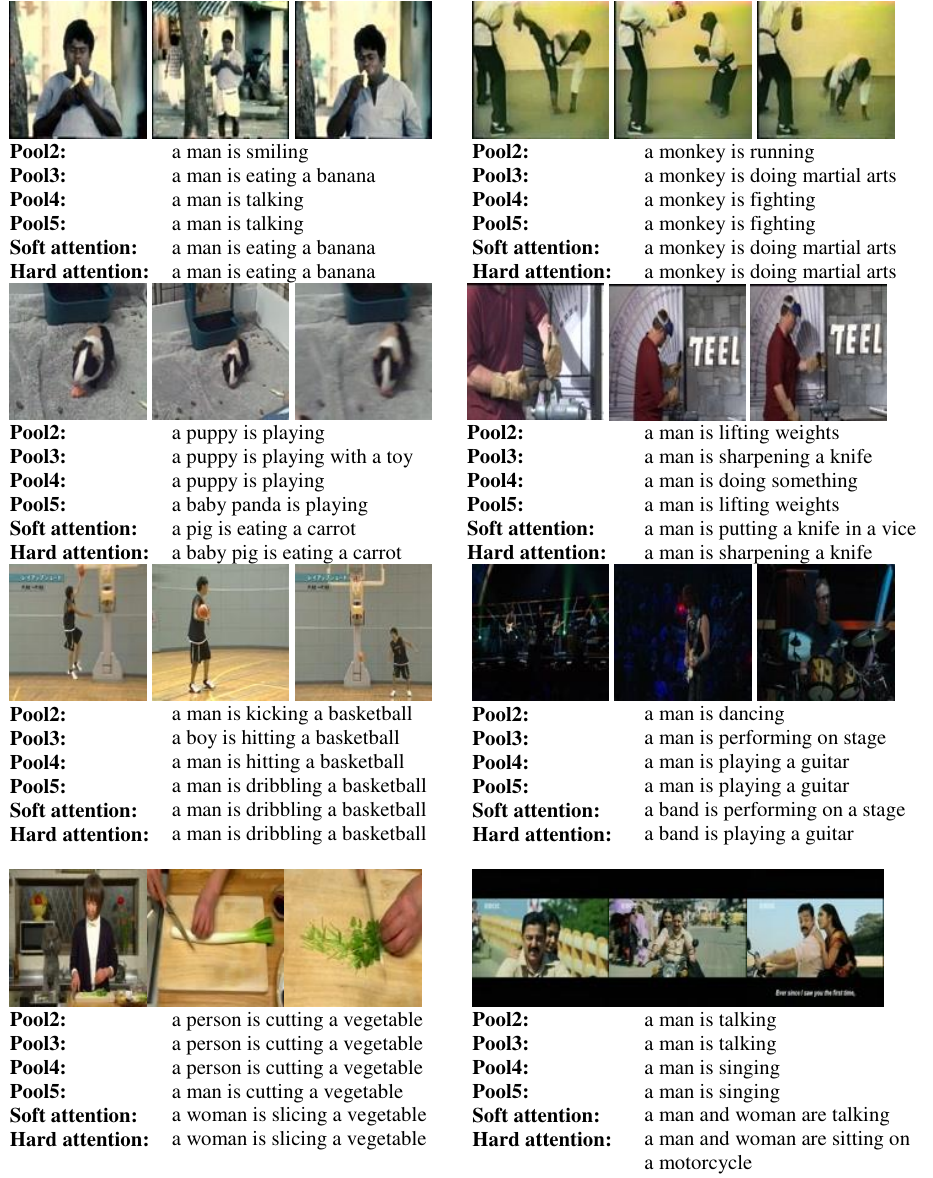}
	\end{center}
}]

\section{Fully Connected Layer Features}
The convolutional-layer features, $\{\av_1^{(n)},\dots,\av_L^{(n)}\}$, are extracted by feeding the entire video into C3D, and hence the dimensions of $\{\av_1^{(n)},\dots,\av_L^{(n)}\}$ are dependent on the video length (number of frames). As discussed in main paper, we employ spatiotemporal attention at each layer (and between layers), and therefore it is not required that the sizes of $\{\av_1^{(n)},\dots,\av_L^{(n)}\}$ be the same for all videos. However, the fully connected layer at the top, responsible for $\av_{L+1}^{(n)}$, assumes that the input video is of the same size for all videos (like the 16-frame-length videos in~\citet{tran2014learning}). To account for variable-length videos, we extract features on the video clips, based on a window of length 16 (as in~\cite{tran2014learning}) with an overlap of 8 frames. $\av_{L+1}^{(n)}$ is then produced by mean pooling over these features. The particular form of pooling used here (one could also use max pooling) is less important than the need to make the dimension of the top-layer features the same for feeding into the final fully-connected layer.

\section{Gradient Estimation for Hard Attention}
Recall the lower-bounded
\begin{align} \label{eq:lbound}
	\log p(\Ymat|\Amat) &= \E_{p(\mv|\Amat)}\log p(\Ymat|\mv,\Amat),
\end{align}
where $\mv = \{\mv_t\}_{t=1,\dots,T}$. Inspired by importance sampling, the multi-sample stochastic lower bound has been recently used for latent variable models~\citep{Burda16importance}, defined as
\begin{align}
	\Lcal^K(\Ymat) = \sum_{\mv^{1:K}}p(\mv^{1:K}|\Amat)\Big[\log  \frac{1}{K}\sum_{k=1}^{K}p(\Ymat|\mv^k,\Amat) \Big],\label{eq:bound}
\end{align}
where $\mv^1,\dots,\mv^K$ are independent samples. This lower bound is guaranteed to be tighter with the increase of the number of samples $K$~\citep{Burda16importance}, thus providing a better approximation of the objective function than (\ref{eq:lbound}). As shown in~\cite{mnih2016variational}, the gradient of $\Lcal^K(\Ymat)$ with respect to the model parameters is 
\begin{align}
	\nabla\Lcal^K(\Ymat) = \sum_{\mv^{1:K}}&p(\mv^{1:K}|\Amat)\sum_{k=1}^{K}\Big[L(\mv^{1:K})\nabla\log  p(\mv^k|\Amat)\nonumber\\
	&+\omega_k \nabla p(\Ymat|\mv^k,\Amat) \Big],
\end{align}
where $L(\mv^{1:K}) = \log  \frac{1}{K}\sum_{k=1}^{K}p(\Ymat|\mv^k,\Amat) $ and $\omega_k = \frac{p(\Ymat|\mv^k,\Amat)} {\sum_j p(\Ymat|\mv^j,\Amat)}$.
A variance reduction technique is introduced in~\cite{mnih2016variational} by replacing the above gradient with an unbiased estimator
\begin{align}
	\nabla\Lcal^K(\Ymat) \approx \sum_{k=1}^{K}&\Big[\hat{L}(\mv^{k}|\mv^{-k})\nabla\log  p(\mv^k|\Amat)\nonumber\\
	&+\omega_k \nabla p(\Ymat|\mv^k,\Amat) \Big],\label{eq:gradient}
\end{align} 
where 
\begin{align}
	\hat{L}(\mv^{k}|\mv^{-k}) & = L(\mv^{1:K}) - \log\frac{1}{K}\big(\sum_{j\neq k}p(\Ymat|\mv^j, \Amat)\nonumber\\
	&~~~~~~~~~+ f(\Ymat, \mv^{-k},\Amat)) \big),\\
	f(\Ymat, \mv^{-k},\Amat) &= \exp(\frac{1}{K-1}\sum_{j\neq k} \log p(\Ymat|\mv^j,\Amat) \,.
\end{align}
When learning the model parameters, the lower bound \eqref{eq:bound} is optimized via the gradient approximation in \eqref{eq:gradient}.

\section{Convolutional Transformation for Spatiotemporal Alignment}
The model architecture of C3D and details of convolutional transformation are provided in Figure~\ref{fig:c3d}. The kernel sizes of the convolutional transformation in (3) in the main paper are $7\times 7 \times 7$, $5\times 5 \times 5$ and $3\times 3 \times 3$ for layer {\em pool2, pool3} and {\em pool4} with $3\times 3 \times 3$, $2\times 2 \times 2$ and $1\times 1\times 1$ zero padding, respectively.  $f(\cdot)$ is implemented by ReLU, followed by 3D max-pooling with $8\times 8 \times 8$, $4\times 4 \times 4$ and $2\times 2 \times 2$ ratios.

The dimensions for features extracted from {\em pool2, pool3, pool4} and {\em pool5} are $28\times 28 \times N/2 \times128$, $14\times 14 \times N/4 \times256$, $7\times 7 \times N/8 \times512$ and $4\times 4 \times N/16 \times512$, respectively. $N$ is the number of frames of input video. After the convolutional transformation, the dimensions will be all  $4\times 4 \times N/16 \times512$. 

To prove these features are spatiotemporal aligned, we first provide the receptive field for 3D convolutional layer and 3D pooling layer. Let $\Ymat = \text{3D-Conv} (\Xmat)$, where 3D-Conv is the 3D convolutional layer with kernel size $3\times 3\times 3$. The features indexed by $i = [i_x,i_y, i_z]$ in $\Ymat$ is obtained by convolving a subset of $\Xmat$ indexed by $j = [j_x,j_y, j_z] $ with convolutional kernel, where $j_x \in [i_x -1, i_x, i_x+1]$, $j_y \in [i_y -1, i_y, i_y+1]$ and $j_z \in [i_z -1, i_z, i_z+1]$. Then, we call that the receptive field of $i = [i_x,i_y, i_z]$ in $\Ymat$ is  $[i_x -1, \dots, i_x+1]\times [i_y -1,\dots, i_y+1] \times  [i_z -1, \dots, i_z+1]$ in $\Xmat$. Similarly, if $\Ymat = \text{3D-pooling} (\Xmat)$ with pooling ratio $2\times 2\times 2$, the receptive field of $i = [i_x,i_y, i_z]$ in $\Ymat$ is  $[2i_x -1, 2i_x]\times [2i_y -1, 2i_y] \times  [2i_z -1, 2i_z]$ in $\Xmat$. 

We then provide the receptive field of features $\av_l$ from each layer in the input video in Table \ref{tab: align_1} and receptive field of features after convolutional transformation, $\hat{\av}_l$, in the original feature $\av_l$ in Tabel \ref{tab: align_2}. The features are all indexed by $i = [i_x,i_y, i_z]$. Combining Table \ref{tab: align_1} and Tabel \ref{tab: align_2}, we can find the receptive field of $\hat{\av}_l$ indexed by $i = [i_x,i_y, i_z]$ for all $l$ in the input video are all 
$[32i_x -63, \dots, 32i_x+30]\times [32i_y -63,\dots, 32i_y+30] \times  [16i_z -32, \dots, 16i_z+15]$.

We index the top-left element in the first frame as $[1,1,1]$. Note that the index of receptive field could be negative due to padding.

\section{Training and Experimental Details}
\subsection{Initialization and Training Procedure}
All recurrent matrices in the LSTM are initialized with orthogonal initialization.
We initialize non-recurrent weights from a uniform distribution in $[-0.01,0.01]$ and all the bias terms are initialized to zero.
Word embedding vectors are initialized with the publicly available \emph{word2vec} vectors that were trained on 100 billion words from Google News, which have dimensionality 300.
The embedding vectors of words not present in the pre-trained set are initialized randomly.
The number of hidden units in the LSTM is set as 512 and we use mini-batches of size $32$.
Gradients are clipped if the norm of the parameter vector exceeds 5. The number of samples for multi-sample stochastic lower bound is set to 10.
The Adam algorithm with learning rate 0.0002 is utilized for optimization.
All experiments are implemented in Torch. 

\twocolumn[{%
	\renewcommand\twocolumn[1][]{#1}%
	\section{Model Architecture of C3D}
	\begin{center}
		\centering
		\includegraphics[width = \textwidth]{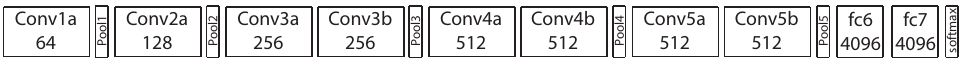}
		\captionof{figure}{\small  C3D net is composed of 8 3D convolution layers, 5 3D max-pooling layers, 2 fully connected layers, and a softmax output layer. All 3D convolution kernels are $3\times 3\times 3$ with $1\times 1 \times 1$ padding and stride 1 in both spatial and temporal dimensions. Number of filters are denoted in each box.
			The 3D max-pooling layers are named from {\em pool1} to {\em pool5}. All pooling  ratios are $2\times 2\times 2$, except for {\em pool1} is $1\times 2\times 2$ (1 is in the temporal dimension).	Each fully connected layer has 4096 output units~\citep{tran2014learning}. }
		\label{fig:c3d}
	\end{center}
	\begin{center}
		\captionof{table}{\small Receptive field of $\av_l$ in the input video.}
		\begin{tabular}{c|l}
			\toprule
			Layer name & Receptive field \\
			\midrule 
			Pool2  & $[4i_x -7, \dots, 4i_x+2]\times [4i_y -7,\dots, 4i_y+2] \times  [2i_z -4, \dots, 2i_z+1]$ \\
			\midrule  
			Pool3 &  $[8i_x -15, \dots, 8i_x+6]\times [8i_y -15,\dots, 8i_y+6] \times  [4i_z -8, \dots, 4i_z+3]$\\
			\midrule 
			Pool4 &  $[16i_x -31, \dots, 16i_x+14]\times [16i_y -31,\dots, 16i_y+14] \times  [8i_z -16, \dots, 8i_z+7]$\\
			\midrule 
			Pool5 &  $[32i_x -63, \dots, 32i_x+30]\times [32i_y -63,\dots, 32i_y+30] \times  [16i_z -32, \dots, 16i_z+15]$\\
			\bottomrule
		\end{tabular}
		\label{tab: align_1}
		
		\captionof{table}{\small Receptive field of $\hat{\av}_l$ in the corresponding $\av_l$.}
		\begin{tabular}{c|l}
			\toprule
			Layer name & Receptive field \\
			\midrule 
			Pool2 &  $[8i_x -14, \dots, 8i_x+7]\times [8i_y -14,\dots, 8i_y+7] \times  [8i_z -14, \dots, 8i_z+7]$\\		
			\midrule 
			Pool3  & $[4i_x - 6, \dots, 4i_x+4]\times [4i_y -6,\dots, 4i_y+4] \times  [4i_z -6, \dots, 4i_z+4]$ \\
			\midrule  
			Pool4 &  $[2i_x -2, \dots, 2i_x+1]\times [2i_y -2,\dots, 2i_y+1] \times  [2i_z -2, \dots, 2i_z+1]$\\
			\bottomrule
		\end{tabular}
		\label{tab: align_2}		
	\end{center}	
	\vspace{6mm}
}]	

\subsection{Max/Average Pooling Baseline Details on Youtube2Text}
A max or average pooling operation is utilized to achieve saptiotemporal alignment, and an MLP is then employed to embed the feature vectors into the same semantic space. 
Specifically, for each $\hat{\av}_l$ with $l = 1,\dots,L$:
\begin{align}
	\tilde{\av}_{i,l}(k) &= \max_{j\in \Ncal_{i,l} } \av_{j,l}(k) \text{  or  } \frac{1}{|\Ncal_{i,l}|}\sum_{j\in\Ncal_{i,l}}  \av_{i,l}(k) \\ 
	\hat{\av}_{i,l}& = \mathsf{MLP}(\tilde{\av}_{i,l})  \,,
\end{align}
where $\Ncal_{i,l}$ is the receptive field (see the previous section for definition) of $\av_{i,L}$ in the $l$-th layer. Similarly, the context vector $\zv_t$ is computed by abstraction-level and spatiotemporal attention.
\end{document}